\documentclass[lettersize,journal]{IEEEtran}
\usepackage{amsmath,amsfonts}
\usepackage{algorithmic}
\usepackage{algorithm}
\usepackage{array}
\pdfoutput=1
\usepackage{textcomp}
\usepackage{stfloats}
\usepackage{url}
\usepackage{verbatim}
\usepackage{graphicx}
\usepackage{cite}

\usepackage{ragged2e}

\begin{document}

\title{Towards Precise Weakly Supervised Object Detection via Interactive Contrastive Learning of Context Information}

\author{Qi Lai,
        Chi-Man Vong,~\IEEEmembership{Senior Member,~IEEE,}

\thanks{Manuscript received December 7, 2022.}}



\maketitle

\begin{abstract}
 Weakly supervised object detection (WSOD) aims at learning precise object detectors with only image-level tags. In spite of intensive research on deep learning (DL) approaches over the past few years, there is still a significant performance gap between WSOD and fully supervised object detection. In fact, most existing WSOD methods only consider the visual appearance of each region proposal but ignore employing the useful context information in the image. To this end, this paper proposes an interactive end-to-end WSDO framework called JLWSOD with two innovations: i) two types of WSOD-specific context information (i.e., \textit{instance-wise correlation} and \textit{semantic-wise correlation}) are proposed and introduced into WSOD framework; ii) an \textit{interactive graph contrastive learning} (iGCL) mechanism is designed to jointly optimize the visual appearance and context information for better WSOD performance. Specifically, the iGCL mechanism takes full advantage of the complementary interpretations of the WSOD, namely instance-wise detection and semantic-wise prediction tasks, forming a more comprehensive solution. Extensive experiments on the widely used PASCAL VOC and MS COCO benchmarks verify the superiority of JLWSOD over alternative state-of-the-art approaches and baseline models (improvement of 3.6\%$\sim$23.3\%  on mAP and  3.4\%$\sim$19.7\% on CorLoc, respectively).
\end{abstract}


\section{Introduction}
\label{sec:intro}

\IEEEPARstart{R}{ecently}, with the advancement of computer vision  object detection, has made great progress \cite{zhao2019object,joseph2021towards,xie2021oriented,fan2021concealed}. Unfortunately, the training for a fully supervised object detection model demands a large number of images with bounding-box annotations (i.e., the labeled ground-truth for the actual objects) which limits the further development of object detection. To alleviate the heavy burden of time-consuming manual labeling procedure to collect bounding-box annotations, weakly supervised object detection (WSOD) \cite{bilen2016weakly,tang2018pcl,zeng2019wsod2,lin2020object,xu2021pyramidal,tang2017multiple,li2019weakly,gao2019c,wan2019c,cinbis2016weakly,girshick2015region} was recently proposed that \textit{only} requires image-level annotations to train object detector. However, the absence of bounding-box level annotations in WSOD causes severe problems, such as region proposal ambiguity and poor-quality bounding box proposals (i.e., the detected object). Consequently, learning object detectors under weak supervision is much more challenging compared to fully supervised scenarios \cite{girshick2015region,ren2015faster,redmon2016you,girshick2015fast}.

\begin{figure*}[!t]
  \centering
   \includegraphics[width=0.8\linewidth]{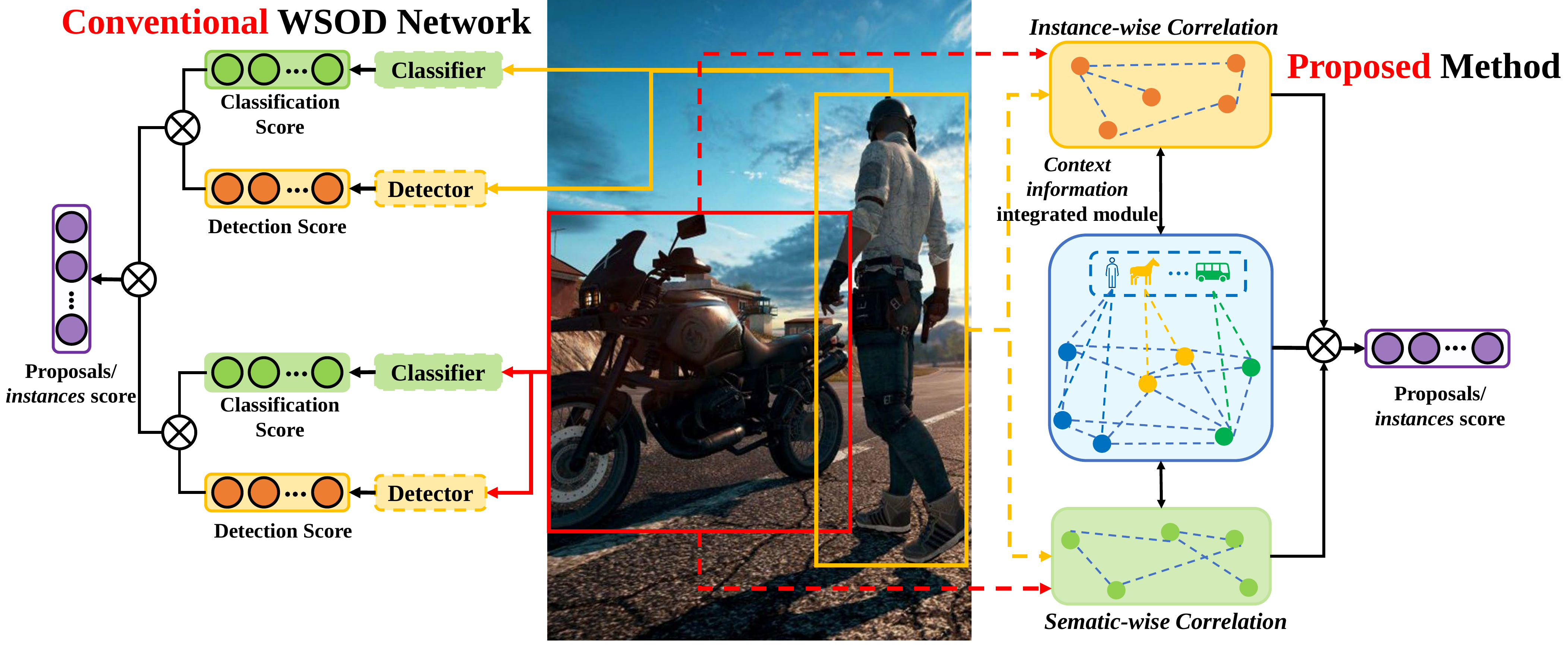}
   \caption{Illustration of the conventional WSOD network (the left part) and our proposed method (the right part).}
   \label{fig:jlwsod}
\end{figure*}

Although existing weakly supervised object detectors can effectively facilitate the WSOD procedure, they do not consider an important information source: \textit{context information}, when inferring the scores of \textit{bounding box regions} (hereafter called \textit{instances} or \textit{proposals}). In fact, under a weakly supervised learning scenarios, the critical challenge is the lack of accurate instance-level annotations. By employing context information, strong predictive capacity can make up for the lack of concrete annotations in weakly supervised learning and therefore increases the robustness in the learning procedure (see Figure \ref{fig:jlwsod} right part).

\IEEEpubidadjcol
Context information is designed to subject to applications. Currently, context information (i.e., joint feature of both local feature and surrounding context) has been successfully applied to different areas such as video analysis \cite{ren2015faster}, image captioning \cite{redmon2016you}, and semantic segmentation \cite{wu2020cgnet} under fully supervised learning. However, such context information for fully supervised learning cannot be easily applied to WSOD task. Therefore, WSOD-specific context information must be designed. 

In WSOD tasks, object localization and semantic ambiguity are two key challenges. In object localization, if precise detection results are necessary to obtain, i.e., localizing the whole object instead of only parts of an object, “local view to global view” correlation and “foreground to background” correlation must be \textit{jointly} taken into consideration. Such \textit{jointly} learned correlation among instances is called \textit{instance-wise correlation}. In semantic ambiguity, if the weakly supervised object detector cannot consider the semantic correlation between each instance, the detector will simply formulate the detection task as a common classification task. So that the detector will assign the wrong category label to the instance within the image. Such semantic correlation between each instance is \textit{semantic-wise correlation}. By introducing the instance-wise correlation and the semantic-wise correlation, the cognitive process and prediction process of visual appearance can be included in the WSOD framework to achieve better detection performance.

For these reasons, this work proposes to introduce these two types of WSOD-specific context information, i.e., \textit{instance-wise correlation} and \textit{semantic-wise correlation}. Specifically, the correlation of the spatially adjacent instances for instance-wise detection (see Figure \ref{fig:jlwsod}) is modeled in order to infer the detection score of each instance. In fact, two significant information can be provided by instance-wise correlation for the detection of  the target object locations: 1) The correlation between the local view (i.e., instances) and the global view (i.e., images); 2) The correlation between the foreground (target object) and the background. By exploring the first correlation, the model can identify the features of some discriminative object parts, that are aggregated into the instances covering the whole object. Meanwhile, with the second correlation, the model can make use of the background to identify the object locations, e.g., a boat should be on the lake. 

To overcome semantic ambiguity, we introduce semantic-wise correlation. It models the intrinsic correlation of co-occurring objects in different categories and facilitates semantic-wise prediction (see Figure \ref{fig:jlwsod}). Semantic-wise correlation helps separate item categories that may occur in an image, particularly under WSOD where the information/features utilized to recognize objects can textitonly be collected through (multi-label) image classification operation \cite{zhang2020weakly}. Consequently, enhancing multi-label image classification by analyzing semantic-wise association may facilitate the acquisition of semantic features for better WSOD performance.

Since these correlations are under weakly supervising scenarios where only image-level labels are available, a self-supervised learning method is necessary. Following the immense success of \textit{contrastive learning} \cite{he2020momentum,grill2020bootstrap,yang2022unified} on self-supervised learning, the graph contrastive learning is employed to learn the two correlations in our WSOD task. However, one technical difficulty arises, contrastive learning can only \textit{independently} learn one type of correlation (i.e., instance-wise correlation or each semantic-wise correlation) while practically these two correlations must be jointly learned. This issue remains unresolved and nontrivial to jointly optimize both correlations.

\begin{figure*}[!t]
\centering
 \includegraphics[width=6in]{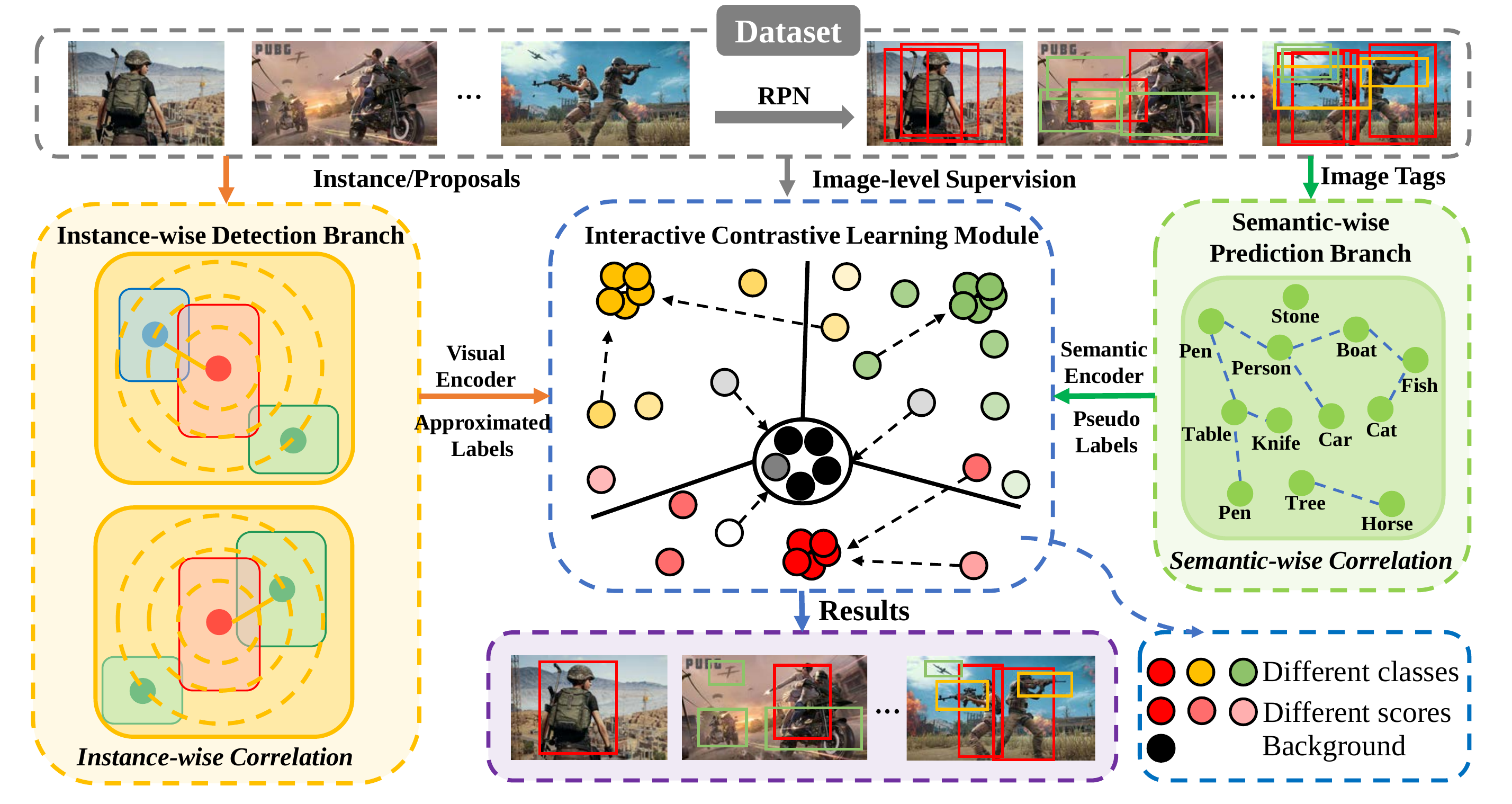}
 \caption{An overview of proposed JLWSOD. Besides the conventional instances/proposal generation and feature extraction, it contains three novel branches to leverage the instance-wise correlation, and semantic-wise correlation to identify the object location and classification, respectively. Notice that the orange arrows indicate the instance/proposal data inputs while the green arrows indicate the semantic information.}
 \label{fig:workfolw}
\end{figure*}

To fill in this blank, a novel interactive end-to-end WSOD framework (Figure \ref{fig:workfolw}) is designed based on interactive contrastive learning of context information. In detail, our WSOD framework is proposed that consists of an \textit{instance-wise detection branch}, a \textit{semantic-wise prediction branch}, and a newly designed \textit{interactive graph contrastive learning} module (iGCL). Since our framework \textit{jointly} leverages instance-wise and semantic-wise information for WSOD, it is called JLWSOD whose workflow is illustrated in Figure \ref{fig:workfolw}. Compared to the existing WSOD frameworks, our JLWSOD can successfully capture different types of context information through the instance-wise detection branch and the semantic-wise prediction branch, which are then jointly optimized through the proposed iGCL interactively. Furthermore, JLWSOD has the ability to conduct inferences at both the instance-level and the semantic-level, which is unexplored in other studies. In summary, the contributions of this paper are as below:

i) Two new WSOD-specific \textit{context information} (i.e., \textit{instance-wise} and \textit{semantic-wise correlation}) are proposed to address the two key challenges in WSOD, namely, object localization and semantic ambiguity, respectively. In this way, our proposed method has inference capacity at the instance-level and semantic-level, which is beyond the exploration of the existing works.

ii) An interactive end-to-end WSOD framework is designed in which the two WSOD-specific context information are separately captured by contrastive learning. Through our newly designed iGCL module, the captured instance-wise and the semantic-wise correlations can be jointly optimized, i.e., complemented by the interactive contrastive learning mechanism so that the object detection performance can be improved, especially on large data sets. This is a non-trivial challenging task because only image-level labels are always available in the training dataset. 

The rest of our article is organized as follows. In Section \ref{sec:Preliminaries}, some preliminaries are introduced. In Section \ref{sec:method}, the details of our method are described. Elaborative experiments and parameters analyze are conducted in Section \ref{sec:exp}. Finally, conclusions are drawn in Section \ref{sec:con}.
\section{Preliminaries}
\label{sec:Preliminaries}
\subsection{Weakly Supervised Object Detection (WSOD)}

Weakly-supervised object detection (WSOD) attracts great interest due to they mainly rely on  image-level annotation, which can alleviate the demand for expensive instance-level annotations. 
For this reason, the WSOD learning framework faces a double challenge. The first challenge is the typical issues (e.g., the intra-class variations in appearance, transformation, scale and aspect ratio) which also encountered in the conventional fully supervised object detection task. The second challenge is the learning efficiency under uncertain challenges caused by the inconsistency between human annotations and real supervisory signals \cite{zhang2021weakly}.

Most existing WSOD methods \cite{bilen2016weakly,tang2018pcl,zeng2019wsod2,xu2021pyramidal} adopt multiple instance learning (MIL) \cite{dietterich1997solving} framework to transform object detection (or \textit{instance} classification) task into image classification (or \textit{bag} classification) task. In MIL, the labeled images are considered as bags and unlabeled image patches (divided from labeled images) as instances. Then, instance classifiers (or object detectors) are trained under MIL constraints, which ensures a positive bag contains at least one positive instance and negative bag contains only negative instances. Currently, MIL framework has become the mainstream for WSOD for its promising results to a certain extent. To further improve WSOD performance, many recent WSOD networks \cite{bilen2016weakly} and their variants \cite{zeng2019wsod2,tang2017multiple,cinbis2016weakly,wan2018min} employ multi-phase network architecture to implement deep MIL detector. In addition, MIL and deep networks are combined by using end-to-end training \cite{diba2017weakly,cheng2022self} to achieve better results for WSOD. 
Due to the absence of instance-level annotations, the performance difference between WSOD and fully supervised object identification remains substantial. For this reason, follow-up works further improve performance through self-training \cite{dietterich1997solving}, knowledge transfer \cite{chen2022lctr,dong2021boosting,hou2021informative}, etc. But they \textit{only} employ one type of input in the training process. In terms of WSOD, the accuracy of object detection and learning procedure are closely related. In the learning procedure, it is crucial to propagate image-level supervisory signals to instance-level (i.e., bounding box-level) training data. Therefore, a framework that can comprehensively consider all useful information in the learning procedure and realize \textit{joint} learning is necessary. This is the research objective of our proposed JLWSOD.
\subsection{Contrastive Learning }

Contrastive learning is a class of self-supervised approaches that train an encoder to be contrastive between the representations that depict statistical dependencies of interest and those that do not. The majority of these approaches use instance discrimination \cite{robert2014machine} as a pretext task to pre-train a network and then fine-tune it for downstream tasks (e.g., classification, object detection, and segmentation). Contrastive learning aims at minimizing the distance between the instances with similar semantics (i.e., positive pairs) and meanwhile stretching away those with divergent semantics (i.e., negative pairs). In particular, Xie et al. \cite{xie2021detco} introduce global view (original image) and local view (instance/proposals) contrastive learning for pretraining and then transfers the trained model to the object detection task. Wang et al. \cite{wang2021multi}  propose a Pseudo Group Contrast (PGC) mechanism to reduce the reliance on pseudo-labels and increase the tolerance for erroneous pseudo-labels in self-training. For better regional features and corresponding semantic descriptions, Yan et al. \cite{yan2022semantics} propose a semantics-guided contrastive network that is beneficial to a variety of downstream tasks. However, all these methods only work on one type of input, such as semantic-wise information or instance-wise information, and cannot do contrastive learning across multiple different information. To alleviate the problem of error amplification caused by inaccurate false labels, and to mine, more unlabeled instances in object detection, an interactive graph contrastive learning module (iGCL) is designed in our JLWSOD. Note that iGCL is applied to incorporate the context information without an extra momentum encoder. Therefore, learning efficiency is kept fast even under an interactive learning mechanism.

\section{methodology}
In this section, the proposed joint learning weakly supervised object detection network (JLWSOD) is elaborated. Firstly, the proposed JLWSOD architecture is illustrated. Then, the formulations of the proposed instance-wise detection branch, semantic-wise prediction branch, and interactive iGCL mechanism are presented. 
\label{sec:method}
\subsection{Network Architecture}

The architecture of our proposed JLWSOD is shown in Figure \ref{fig:workfolw}. The $X$ is an image with corresponding image-level tags $\textbf{Y}=[\textbf{y}_1,\textbf{y}_2,…,\textbf{y}_k]\in \mathbb{R}^K$, where $\textbf{y}_k\in [0,1]$ and $K$ is the number of target semantic categories. The image $X$ first passes through a backbone network, e.g., ResNet50 \cite{he2016deep} to extract the feature maps. Here, a set of instances or a \textit{bag} $B=\left \{\ {b}_i\right \}_{i=1}^{\left |B \right |} $  are generated using any proposal generation methods \cite{zitnick2014edge,pont2016multiscale,kong2016hypernet} (e.g., HyperNet \cite{kong2016hypernet} here). And the $\left | \cdot  \right |$ denotes the number of elements in a set. Then, ROI pooling for each instance $b_i$ is performed to obtain its corresponding feature vector $\textbf{x}_{i}$.

After obtaining the deep feature, there are two parallel network branches, i.e., the instance-wise detection branch  $\mathbb{B}_{ins}$ and the semantic-wise prediction branch  $\mathbb{B}_{sem}$, which share the same architecture but with different sets of weights. These two branches can recall potential objects respectively based on image features and sematic-wise correlation between the labeled image and the unlabeled instance proposals. 

Nonetheless, if trained a model is only based on the pseudo labels generated by the above two branches, the model is easily be confused by the false pseudo labels. Because it focuses on independently learning a type of input (either instance or semantic) and heavily relies on the image-level tags for class discrimination. Therefore, we further design an interactive graph contrastive learning module (iGCL) for context information which interactively boosts the supervision of each branch under a weakly supervised setting. 

This idea is inspired by the fact that hybrid contrastive learning loss focuses on jointly adopting context information and is naturally independent of false pseudo labels.

Finally, the overall network is trained by jointly optimizing the following composite loss function:
\begin{equation}
  \mathcal{L} = \lambda_{ins}\mathcal{L}_{ins}+\lambda_{sem}\mathcal{L}_{sem}+\lambda_{iGCL}\mathcal{L}_{iGCL}
  \label{eq:total_loss}
\end{equation}
where $\mathcal{L}_{ins}$ and $\mathcal{L}_{sem}$ are the loss functions of the instance-wise detection branch and the semantic-wise prediction branch, respectively. The $\mathcal{L}_{iGCL}$ is the contrastive loss of iGCL. $\lambda_{ins}$, $\lambda_{sem}$, $\lambda_{iGCL}>0$ are tuning parameters to weigh the importance of $\mathcal{L}_{ins}$, $\mathcal{L}_{sem}$ and $\mathcal{L}_{iGCL}$, respectively. The above composite loss function can be optimized in an end-to-end manner. 

The above composite loss function can be optimized in an end-to-end manner. For clarity, the detailed description of our JLWSOD is provided in Algorithm \ref{alg:alg1}. In the test stage, only the projector $f_{\phi _{ins}}$ and a predictor $f_{\phi _{sem}}$ are needed so that our method as efficient as most existing methods [9-11, 39], the detail as illustrated in Section \ref{sec:con}.

\begin{algorithm}[H]
\caption{JLWSOD}\label{alg:alg1}
\begin{algorithmic}
\STATE \textbf{Input}: feature vector $\textbf{x}$, image-level tags $\textbf{Y}$, maximum number of iterations of instance-wise detection $T_{ins}$, maximum number of iterations of prediction $T_{sem}$, maximum iterations of contrastive learning $T_{con}$, number of categories $K$.
\STATE \textbf{Output}: predicted result $\hat{\textbf{O}}$.
\STATE 1: \textbf{Steps of approximated label generation:}
\STATE 2: \textbf{For} $\textbf{x}$, $t_{sem}$, $\textbf{Y}$ in loader:
\STATE 3: \hspace{0.5cm} Calculate the detection loss $\mathcal{L}_{ins}$ according to Eq. (\ref{eq:l_ins});
\STATE 4: \hspace{0.5cm} Obtain $\tilde{y}_i^k$ according to Eq. (\ref{eq:det_y});
\STATE 5: \textbf{End for}
\STATE 6: \textbf{Steps of pseudo label generation:}
\STATE 7: \hspace{0.5cm} for $\textbf{x}$, $t_{sem}$, $\textbf{Y}$ in loader:
\STATE 8: \hspace{0.5cm} Calculate prediction loss $\mathcal{L}_{sem}$ according to Eq. (\ref{eq:y_sem});
\STATE 9: \hspace{0.5cm} Obtain the pseudo label $\hat{\mathbf{y}}_i$ according to Eq. (\ref{eq:y_sem});
\STATE 10: \textbf{End for} 
\STATE 11: \textbf{Steps of interactive graph contrastive learning:}
\STATE 12: \textbf{For} $\textbf{x}$, $\textbf{z}_i$, $t_{con}$, $\tilde{y}_i^k$, $\hat{\mathbf{y}}_i$ in loader:
\STATE 13: \hspace{0.5cm} Initialize $f_{\phi _{ins}}$ and $f_{\phi _{sem}}$;
\STATE 14: \hspace{0.5cm} Calculate the latent embedding $\textbf{u}_i$, ${\textbf{u}_i}'$, $\textbf{v}_i$, ${\textbf{v}_i}'$ according to Eqs. (\ref{eq:ins_iGCL}) and (\ref{eq:sem_iGCL});
\STATE 15: \hspace{0.5cm} Calculate interactive graph contrastive loss $\mathcal{L}_{iGCL}$ based on Eq. (\ref{eq:l_iGCL});
\STATE 16: \textbf{End for} 

\end{algorithmic}
\label{alg1}
\end{algorithm}

\subsection{Instance-wise detection branch}

To detect the locations of potential objects, most existing methods follow WSDDN \cite{bilen2016weakly} that uses a detection branch to compute the detection scores of each instance $b_i$ by using a cross-instance SoftMax operator, after a learner mapping function $f_{det}$. The $f_{det}$ is defined as
\begin{equation}
  \varphi_i=f_{det}\left (\mathbf{x}_i,\mathbf{w}_{det} \right)
  \label{eq:detection}
\end{equation}
where $\textbf{x}_i$ is the feature vector of the instance $b_i$. $\varphi_i$ and $\mathbf{w}_{det}$ are the detection scores and the learner parameters in $f_{det}\left ( \cdot  \right ) $. Obviously, the detection scores computed by Eq. (\ref{eq:detection}) depend only on the visual information of each instance, ignoring useful context information. Under weakly supervised learning scenarios where concrete bounding box annotations are missing during the training procedure, this formulation would lead existing methods to acquire undesired instance regions that focus on small discriminative object parts rather than the whole object.

To tackle this problem, a novel \textit{instance-wise detection branch} $\mathbb{B}_{ins}$ is devised to exploit the correlation of the spatially adjacent instances. The design of $\mathbb{B}_{ins}$ follows Faster R-CNN \cite{ren2015faster}. In detail, $\mathbb{B}_{ins}$ takes feature vector $\textbf{x}_i$ and a bag $B$ as input, through the function $f_{ins}$ computes the probability of each instance $b_i \in B $ belonging to each class $(k = 1$ to $K)$ under the supervision of image-level tags, and formulated as below:
\begin{equation}
  corr_{ins}=f_{ins}\left (\mathbf{x}_i,B \right)
  \label{eq:detection}
\end{equation}
where $corr_{ins} \in \left [ 0,1 \right ]^{\left | B \right |\times  K} $ and each element $corr_{ins} (i,k)$ indicates the probability of the $i$-th instance $b_i$ belonging to $k$-th category. Suppose the aggregation score of all instances for each category $k$ in the image is $\mathcal{S}(:,k)$, which can be inferred from $corr_{ins} (i,k)$. Different from the Eq. (\ref{eq:detection}), Log-Sum-Exp (LSE) function \cite{robert2014machine} is used to calculate a smooth approximation to the maximum value of $corr_{ins}(i,k)$  $(i=1,2,…,\left | B \right |)$, which can be denoted as 
\begin{equation}
  \mathcal{S}=\frac{1}{r} log\left [ \frac{1}{\left | B \right|} \sum_{i=1}^{\left | B \right|} \exp (r(corr_{ins} (i,k))) \right ] 
  \label{eq:lse}
\end{equation}
where $r$ is the parameter allonging LSE function to behave in a range between the maximum and the average. Compared with a simple maximum, LSE function approximates the maximum and considers all elements of $corr_{ins} (i,k)$. With the smoothly approximated $\mathcal{S}(i,k)$, the instance-wise detection loss function is defined as
\begin{equation}
  \tilde{y}_i^k=\rho (corr_{ins} (i,k)) 
  \label{eq:det_y}
\end{equation}
\begin{equation} 
\label{eq:l_ins}
\begin{aligned}
\mathcal{L}_{ins}=&\sum_{k=1}^{K} \mathcal{L}_{MCE}\left [\sum_{i=1}^{\left | B \right|} corr_{ins} (i,k), y_k \right ]\\
&+\sum_{k=1}^{K+1}\sum_{i=1}^{\left | B \right|} \mathcal{L}_{CE}\left [\mathcal{S}(i,k), \tilde{y}_i^k \right ] 
\end{aligned}
\end{equation}
where $\tilde{y}_i^k$ indicates the approximated category label for $i$-th instance belong to $k$-th class. $\rho(\cdot )$ is the indicator function. $\mathcal{L}_{MCE}$ and $ \mathcal{L}_{CE}$ are the standard multi-class cross-entropy (MCE) loss and the weighted cross entropy (CE) loss \cite{tang2017multiple}. Here, $ \mathcal{L}_{CE}$ is adopted to replace $ \mathcal{L}_{MCE}$  because each instance has one and only one positive category label.

\subsection{Semantic-wise prediction branch}

To obtain the final classification score vector of each training image, previous WSOD methods usually directly perform the multiplication-and-addition operation, i.e., conducting element-wise multiplication on the score vectors obtained from the detection and the classification branches, and then adding the obtained score vectors across the instances within the given image, as illustrated in the left part of Figure \ref{fig:workfolw}). However, converting the instance-level score vectors to bag/image-level score vectors in this way, \textit{semantic correlation} is ignored, resulting in inaccurate image-level score vectors, especially when images contain objects from multiple categories.
However, transiting the instance-level score vectors to the bag/image-level score vectors in such a way, the \textit{semantic-wise correlation} is ignored, leading to inaccurate image-level score vectors, especially when an image (bag) contains multiple categories of objects/instances.

To solve this problem, a \textit{semantic-wise prediction branch} $\mathbb{B}_{sem}$ is proposed to mine all latent object categories that appear in each training image. Rather than simply employing the multiplication-and-addition operation, the proposed $\mathbb{B}_{sem}$ additionally makes use of \textit{semantic-wise correlation}. Consequently, the visual information and the semantic-wise correlation must be considered together, and a semantic space is constructed, on which the visual feature \textbf{x} is projected. In particular, $\textbf{x}_i$ refers to the feature vector of $i$-th instance of an image, generated by the proposal generation methods. Then $\textbf{x}_i$ is transformed into a d-dimensional vector $\textbf{z}_i$ through pre-trained deep semantic segmentation networks (e.g., ResNet50 \cite{he2016deep}) and formulated as

\begin{equation}
  \mathbf{z}_i=f_{sem} \left (\mathbf{w}_{i}^{sem},\mathbf{x}_i \right)
  \label{eq:z_i}
\end{equation}
where $\mathbf{w}_{i}^{sem}\in \mathbb{R}^{d\times D}$  represents the weight matrix of semantic segmentation.

Subsequently, the \textit{semantic covariance }of bag-of-instances is defined as follows. Suppose $\mathbf{w}_{i}^{sem}$ denotes the matrix formed by treating each vector $\mathbf{w}_{i}^{(q)}$, $q=1$ to $D$ as a column, and let $\mathbf{w}_{i}^{(p)}$, $\mathbf{w}_{i}^{(p,q)}$ denote a row vector and an element of this matrix, respectively. The \textit{semantic covariance} of $i$-th instance and $j$-th instance ($i,j=1,2,…,\left | B \right|$ and $i\ne j$) is defined as: 
\begin{equation} 
\label{eq:cov_s}
\begin{aligned}
cov_s(\mathbf{z}_i,\mathbf{z}_j)&=\frac{1}{D}\sum_{q=1}^{D}((\mathbf{w}_{i}^{(p,q)}-\bar{\mathbf{w}}_i^p)(\mathbf{w}_{j}^{(p,q)}-\bar{\mathbf{w}}_j^p))\\
&=\frac{1}{D}\sum_{q=1}^{D}(\mathbf{w}_{i}^{(p,q)}\mathbf{w}_{j}^{(p,q)}-\bar{\mathbf{w}}_i^p\bar{\mathbf{w}}_j^p)
\end{aligned}
\end{equation}
where $\bar{\mathbf{w}}_j^p$ means the $p$-th row in $\mathbf{w}_{j}^{sem}$ and $\mathbf{w}^{sem}=\left \{ \mathbf{w}_{i}^{sem} \right \}_{i=1}^{\left | B \right|}$. Then, the \textit{semantic-wise correlation} is:

\begin{equation}
 corr_{sem}(\mathbf{z}_i,\mathbf{z}_j)=\frac{cov_s(\mathbf{z}_i,\mathbf{z}_j)}{\sqrt{cov_s(\mathbf{z}_i,\mathbf{z}_i)cov_s(\mathbf{z}_j,\mathbf{z}_j)}} 
  \label{eq:corr_sem}
\end{equation}

After obtaining the $corr_{sem}$ which is a $d\times d$ matrix, the pseudo category label $\hat{\mathbf{y}}_i$ for each instance can be generated:
\begin{equation}
 \hat{\mathbf{y}}_i=corr_{sem}\mathbf{w}_{i}^{sem}\mathbf{x}_i
  \label{eq:y_sem}
\end{equation}
where $\hat{\mathbf{y}}_i$ is the output of the $\mathbb{B}_{sem}$ and $\hat{y}_i=arg\max_{i}(\hat{\mathbf{y}}_i) $. To this end, the loss function of $\mathbb{B}_{sem}$ is introduced to aggregate semantic features with similar semantics meanings.
\begin{equation}
 \mathcal{L}_{sem}= \frac{1}{\left | B \right|} \sum_{i=1}^{\left | B \right|} \left [ 1-\frac{\mathbf{z}_i \cdot \mathbf{c}_{\hat{y}_i}}{\left \| \mathbf{z}_i  \right \|_2 \left \| \mathbf{c}_{\hat{y}_i} \right \|_2 }  \right ] 
  \label{eq:l_sem}
\end{equation}
where $\mathbf{c}_{\hat{y}_i}$ denotes the learning center for the category of $i$-th instance, and $\left \| \cdot \right \|_2 $ is the $l_2$-norm for a vector. In each training iteration, $\mathbf{c}_{\hat{y}_i}$ is updated relative to semantic feature vector $\mathbf{z}_i$ as
\begin{equation}
 \mathbf{c}_{\hat{y}_i}^{t+1}=\mathbf{c}_{\hat{y}_i}^{t}+\theta  \cdot (\mathbf{z}_i-\mathbf{c}_{\hat{y}_i}^{t})
  \label{eq:c_y}
\end{equation}
where $\theta$ means the learning rate. As a result, the similarity distance between instance pairs can be calculated by their learned semantic feature vector $\mathbf{z}_i$ by using Eq. (\ref{eq:c_y}). 


\subsection{Interactive Contrastive Learning Mechanism}

Since the lack of instance-level annotations (ground truth), network optimization in WSOD cannot be done by calculating the distance between the predicted results and ground truths. Inspired by \cite{zheng2021generative}, an \textit{interactive graph contrastive learning} (iGCL) is designed to complement both detection and segmentation for more accurate prediction.

\begin{figure}[ht]
  \centering
   \includegraphics[width=1\linewidth]{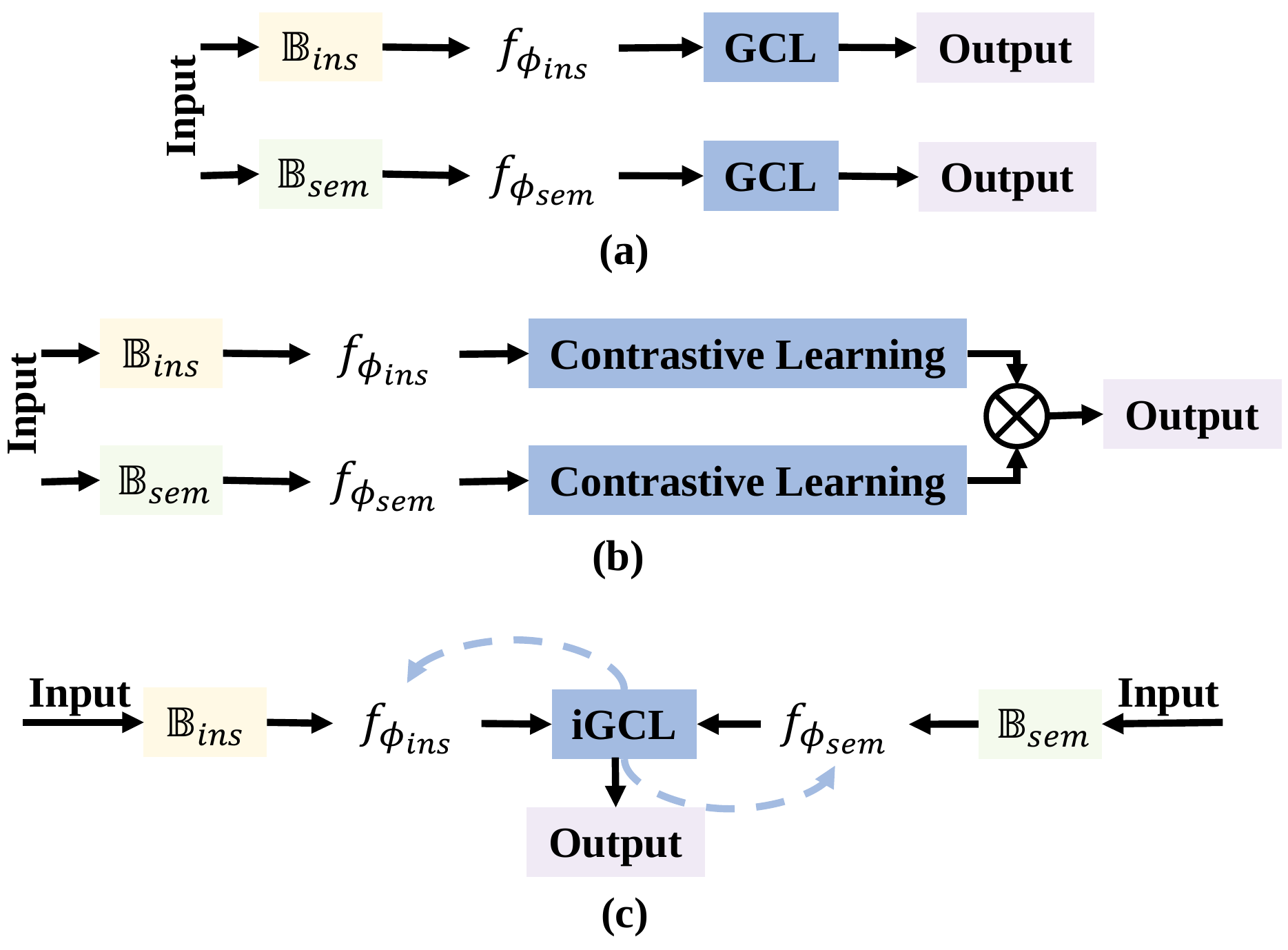}
   \caption{Illustration of the interactive graph contrastive learning mechanism. (a) is a standard \textit{graph contrastive learning} (GCL) workflow, which only considers the single input and learning contrast from one perspective: either instance-level or semantic-level. (b) is a simple multiplication-and-addition operation of two independent GCLs. (c) is our proposed \textit{interactive graph contrastive learning} (iGCL) framework, which can learn the representation interactively, building contrastive loss across the instance-level and semantic-level information.}
   \label{fig:iGCL}
\end{figure}

As we discussed above, the instance-wise detection branch $\mathbb{B}_{ins}$ is very easy to over-fit to the discriminative parts of an object so that only parts of the object region can be detected/covered. Meanwhile, the semantic-wise prediction branch $\mathbb{B}_{sem}$ can successfully cover the whole object region. Intuitively, $\mathbb{B}_{sem}$ can be used to refine the results of $\mathbb{B}_{ins}$ by using semantic-wise correlation to filter proposals with larger IoU. On the other hand, since $\mathbb{B}_{ins}$ is able to distinguish an object from the background effectively, $\mathbb{B}_{ins}$ can guide $\mathbb{B}_{sem}$ to generate a more accurate pseudo-category label. Therefore, an interactive learning objective between instance-level and semantic-level is considered:
\begin{equation}
 \mathcal{L}_{iGCL}=\mathcal{L}_{con(S\to D)}+\mathcal{L}_{con(D\to S)}
  \label{eq:l_iGCL}
\end{equation}
where the loss of iGCL has two parts: the $\mathbb{B}_{sem} \to \mathbb{B}_{ins}$ contrastive loss $\mathcal{L}_{con(S\to D)}$  and the $\mathbb{B}_{ins} \to \mathbb{B}_{sem}$ contrastive loss $\mathcal{L}_{con(D\to S)}$. We follow BYOL \cite{grill2020bootstrap} for learning contrastive representation. The $\mathbb{B}_{ins}$ and the $\mathbb{B}_{sem}$ are appended with projectors $f_{\phi _{ins}}$, $f_{\phi_{ins}}'$ and predictors $f_{\phi _{sem}}$, $f_{\phi _{sem}}'$ for obtaining latent embeddings. Both $f_{\phi _{ins}}$ and $f_{\phi _{sem}}$ are two-layered GCNs. $\textbf{u}_i$, ${\textbf{u}_i}'$, $\textbf{v}_i$ and $\textbf{v}_i'$ are employed to indicate the latent embedding of $\textbf{x}_i$, $\tilde{\textbf{y}}_i$, $\textbf{z}_i$ and $\hat{\mathbf{y}}_i$, respectively.
\begin{equation}
\mathbf{u}_i=f_{\phi _{ins}}(\mathbf{x}_i);{\mathbf{u}_i}'={f_{\phi _{ins}}}'(\tilde{\textbf{y}}_i)
  \label{eq:ins_iGCL}
\end{equation}
\begin{equation}
\mathbf{v}_i=f_{\phi _{sem}}(\mathbf{v}_i);{\mathbf{v}_i}'={f_{\phi _{sem}}}'(\hat{\textbf{y}}_i)
  \label{eq:sem_iGCL}
\end{equation}
where the $\tilde{\textbf{y}}_i={\left [\tilde{y}_i^1,\tilde{y}_i^2,...,\tilde{y}_i^K \right ] }$ is obtained by Eq. (\ref{eq:det_y}). The $\textbf{z}_i$ and $\hat{\mathbf{y}}_i$ obtain by Eqs. (\ref{eq:z_i}) and (\ref{eq:det_y}), respectively. Then, the $\mathbb{B}_{sem} \to \mathbb{B}_{ins}$ contrastive loss $\mathcal{L}_{con(S\to D)}$ can be formulated as:
\begin{equation}
\mathcal{L}_{con(S\to D)}=-\sum_{i\in \left | B \right | } \frac{1}{\left | B \right |} log\frac{exp(\tau \mathbf{u}_i^T \mathbf{v}_i)}{\sum_{j\in \left | B \right |} exp(\tau \mathbf{u}_i^T \mathbf{v}_j)} 
  \label{eq:l_sd}
\end{equation}

Similarly, the $\mathbb{B}_{ins} \to \mathbb{B}_{sem}$ contrastive loss $\mathcal{L}_{con(D\to S)}$ is
\begin{equation}
\mathcal{L}_{con(D\to S)}=-\sum_{i\in \left | B \right | } \frac{1}{\left | B \right |} log\frac{exp(\tau {{\mathbf{u}}'}_i^T {\mathbf{v}_i}')}{\sum_{j\in \left | B \right |} exp(\tau {{\mathbf{u}}'}_i^T {\mathbf{v}}'_j)} 
  \label{eq:l_ds}
\end{equation}
where $\left | B \right |$ is the number of instances. By learning the representation between the instance-level and semantic-level, instance recognition at the semantic-level and instance-level are mutually enhanced. For this reason, detection and classification performance can be boosted. The detail of our proposed iGCL is illustrated in Figure \ref{fig:iGCL}.

\section{Experiments}
\label{sec:exp}

\subsection{Datasets and Evaluation Metrics}

Following recent literature \cite{grill2020bootstrap,dong2021boosting,hou2021informative,chen2021points,li2022siod}, PASCAL VOC 2007 \cite{everingham2015pascal}, MS-COCO 2014 and MS-COCO 2017 \cite{lin2014microsoft} datasets were employed to evaluate the performance of JLWSOD. Note that the experiments only employed image-level tags for training. VOC2007 dataset consists of 20 semantic categories and 9,943 images. Both  correct localization (CorLoc) and mean average precision (mAP) are taken as the evaluation metrics for the  VOC2007 dataset. CorLoc is adopted to measure the object localization performance on the weakly supervised training samples, while mAP measures the object detection performance on the test samples. Both these two metrics follow the standard PASCAL VOC protocol, i.e., intersection-over-union (IoU) $>0.5$ between the detected boxes with the ground truth ones. The MS-COCO datasets are much larger than the Pascal VOC dataset which contains 80 object categories. We follow \cite{bilen2016weakly,tang2018pcl,zeng2019wsod2} to train on the standard training set (41K images) and validation set (5K images) while evaluating the test set (about 20K images). The evaluation metrics mAP@0.5 and mAP@$\left [ 0.5:0.05:0.95 \right ] $ based on standard MS-COCO criteria are employed, i.e., computing the average of mAP for IoU$\in[0.5:0.05:0.95]$.

\subsection{Implementation Details}

Our method is based on ResNet50 \cite{he2016deep}, which is pre-trained on the training set of each dataset. To reduce the number of inaccurate instances, we filter out the instances whose width or height is smaller than 16 pixels. During training, the mini-batch size for SGD is respectively set to be 4, and 32 for VOC 2007, and MS-COCO. An initial learning rate is set to 1e-3 for the first 40, 15, and 85K iterations for the VOC 2007 and MS-COCO datasets, respectively. Then the learning rate is decreased to 1e-4 in the following 10, and 20K iterations for the VOC 2007 and MS-COCO datasets, respectively. The momentum and weight decay are set to be 0.9 and 0.0005 respectively. All our experiments are run on NVIDIA GTX 1080Ti GPU with 16 RAM memory.


\subsection{Results and Comparisons}

 \begin{table*}[]
 \caption{Result (mAP in \%) for compared methods on the VOC 2007 Test Set.}
\label{tab:map}
\resizebox{\linewidth}{!}{
\begin{tabular}{lccccccccccccccccccccc}
\hline
Method                         & Aero & Bike & Bird & Boat & Bottle & Bus  & Car  & Cat  & Chair & Cow  & Table & Dog  & Horse & Mbike & Person & Plant & Sheep & Sofa & Train & tv   & mAP  \\ \hline
WSSDN \cite{bilen2016weakly}                  & 39.2 & 48.4 & 32.6 & 17.2 & 11.8   & 64.2 & 47.7 & 42.2 & 10.2  & 37.4 & 24.8  & 38.2 & 34.4  & 55.6  & 9.4    & 15.0  & 32.1  & 42.7 & 54.7  & 46.9 & 35.2 \\
PCL \cite{tang2018pcl}                    & 54.5 & 69.1 & 39.2 & 19.2 & 15.2   & 64.2 & 63.8 & 30.1 & 25    & 52.5 & 44.3  & 19.5 & 39.6  & 67.8  & 17.5   & 23.9  & 46.7  & 57.5 & 58.6  & 62.8 & 43.5 \\
WCCN \cite{diba2017weakly}                    & 49.5 & 60.6 & 38.6 & 29.2 & 16.2   & 70.8 & 56.9 & 42.5 & 10.9  & 44.1 & 29.9  &42.2  & 47.9  &64.1   &13.8    &23.5   &45.9   &54.1  &60.8  &54.5  &42.8  \\
WPRN \cite{tang2018weakly}           &57.9  &70.5  &37.8 & 5.7 &21.0    &66.1  &\textbf{69.2}  &59.4  &3.4     &57.1  &\textbf{57.3}   &35.2  &64.2   &68.6   &32.8    &28.6   &50.8   &49.5  &41.1  &30.0  &45.3  \\
MELM \cite{wan2018min}           &55.6  &66.9  &34.2 &29.1  &16.4   &68.8  &68.1  &43.0  &25.0     &\textbf{65.6}  &45.3   &53.2  &49.6   &68.6   &2.0    &25.4   &52.5   &56.8  & 62.1 &57.1  &47.3  \\
OICR \cite{tang2017multiple}                  & 58.1 & 62   & 31.8 & 19.1 & 12.8   & 65.5 & 62.5 & 28.3 & 24.4  & 45.0 & 30.6  & 25.3 & 37.9  & 65.6  & 15.5   & 24    & 41.8  & 47.0 & 64.6  & 62.6 & 41.2 \\
C-MIL \cite{wan2019c}                 & 62.5 & 58.4 & 49.5 & 32   & 19.8   & \textbf{70.8} & 66.2 & 63.5 & 20.5  & 60.7 & \textbf{51.6}  & 53.8 & 57.7  & \textbf{69.1}  & 8.4    & 23.6  & 51.8  & 58.5 & \textbf{66.9}  & 63.7 & 50.5 \\
SCDN \cite{li2019weakly}                  & 59.4 & \textbf{71.5} & 38.9 & 32.2 & 21.5   & 66.7 & 64.5 & \textbf{68.9} & 20.4  & 49.2 & 47.6  & 60.9 & 55.9  & 67.7  & 30.2   & 22.9  & 45    & 53.2 & 60.9  & 64.5 & 50.1 \\
WSOD$^2$ \cite{zeng2019wsod2}                  & 65.1 & 64.9 & 54.9 & 39.6 & 22.2   & 69.8 & 62.8 & 61.4 & 29.1  & 64.4 & 42.9  & 60.2 & \textbf{70.1}  & 68.7  & 18.9   & 24.1  & \textbf{58.6}  & \textbf{59.7} & 55.5  & \textbf{64.9} & 52.9 \\
MIST \cite{ren2020instance}                  & —    & —    & —    & —    & —      & —    & —    & —    & —     & —    & —     & —    & —     & —     & —      & —     & —     & —    & —     & —    & 54.9 \\
P-MIDN \cite{xu2021pyramidal}                 & —    & —    & —    & —    & —      & —    & —    & —    & —     & —    & —     & —    & —     & —     & —      & —     & —     & —    & —     & —    & 53.9 \\
\textit{Zhang} \cite{he2020momentum} & 62.2 & 61.1 & 51.1 & 33.8 & 18.1   & 66.7 & 66.5 & 64.9 & 18.5  & 59.4 & 44.8  & 60.9 & 65.6  & 66.9  & 24.7   & 25.9  & 51.1  & 53.2 & 65.9  & 62.2 & 51.2 \\
\textbf{Ours}                  & \textbf{67.7} & 68.5 & \textbf{58.2}& \textbf{45.1} & \textbf{43.3}   & 69.3 & 65.2 & 65.1 & \textbf{40.4}  & 61.7 & 50.2  & \textbf{63.3} & 67.3  & 66.1  & \textbf{45.5}   & \textbf{46.2}  & 58.1  & 57.4 & 65.9  & 64.1 & \textbf{58.5} \\ \hline
\multicolumn{20}{l}{\textit{Remark: “—” denotes not reported in the literature.}}\\
\end{tabular}
}
\end{table*}

\begin{table*}[]
\caption{Result (CorLoc in \%) for compared methods on the VOC 2007 Train and Validation Set}
  \label{tab:corloc}
\resizebox{\linewidth}{!}{
\begin{tabular}{lccccccccccccccccccccc}
\hline
Method                         & Aero & Bike & Bird & Boat & Bottle & Bus  & Car  & Cat  & Chair & Cow  & Table & Dog  & Horse & Mbike & Person & Plant & Sheep & Sofa & Train & tv   & CorLoc \\ \hline
WSSDN \cite{bilen2016weakly}                  & 65.1 & 58.8 & 58.5 & 33.1 & 39.8   & 68.3 & 60.2 & 59.6 & 34.8  & 64.5 & 30.5  & 42.9 & 56.8  & 82.4  & 25.5   & 41.6  & 61.5  & 55.9 & 65.9  & 63.7 & 53.5   \\
PCL \cite{tang2018pcl}                    & 79.6 & \textbf{85.5} & 62.2 & 47.9 & 37.0   & 83.8 & 83.4 & 43.0 & 38.3  & 80.1 & 50.6  & 30.9 & 57.8  & \textbf{90.8}  & 27.0   & 58.2  & 75.3  & \textbf{68.5} & 75.7  & 78.9 & 62.7   \\
WCCN \cite{diba2017weakly}                  &83.9  &72.8  &64.5  &44.1  &40.1   &65.7  &82.5  &58.9 &33.7 &72.5  &25.6 &53.7  &67.4 &77.4 &26.8&49.1&68.1 &27.9  &64.5  &55.7  &56.7  \\
WPRN \cite{tang2018weakly}           &77.5  &81.2 &55.3 &19.7  &44.3 &80.2  &86.6  &69.5  &10.1 &87.7 &68.4 &52.1 &84.4 &91.6  &57.4 &63.4 &77.3 &58.1 &57.0 &53.8 &63.8  \\
MELM \cite{wan2018min}         & —    & —    & —    & —    & —      & —    & —    & —    & —     & —    & —     & —    & —     & —     & —      & —     & —     & —    & —     & —    & —      \\
OICR \cite{tang2017multiple}                  & 81.7 & 80.4 & 48.7 & 49.5 & 32.8   & 81.7 & \textbf{85.4} & 40.1 & 40.6  & 79.5 & 35.7  & 33.7 & 60.5  & 88.8  & 21.8   & 57.9  & 76.3  & 59.9 & 75.3  & \textbf{81.4} & 60.6   \\
C-MIL \cite{wan2019c}                 & —    & —    & —    & —    & —      & —    & —    & —    & —     & —    & —     & —    & —     & —     & —      & —     & —     & —    & —     & —    & —      \\
SCDN \cite{li2019weakly}                  & 85   & 83.9 & 58.9 & 59.6 & 43.1   & 79.7 & 85.2 & \textbf{77.9} & 31.3  & 78.1 & 50.6  & 75.6 & 76.2  & 88.4  & 49.7   & 56.4  & 73.2  & 62.6 & 77.2  & 79.9 & 68.6   \\
WSOD$^2$ \cite{zeng2019wsod2}                  & \textbf{87.1} & 80   & \textbf{74.8} & 60.1 & 36.6   & 79.2 & 83.8 & 70.6 & 43.5  & \textbf{88.4} & 45.9  & 74.7 & \textbf{87.4}  & \textbf{90.8}  & 44.2   & 52.4  & \textbf{81.4}  & 61.8 & 67.7  & 79.9 & 69.5   \\
MIST \cite{ren2020instance}                  & —    & —    & —    & —    & —      & —    & —    & —    & —     & —    & —     & —    & —     & —     & —      & —     & —     & —    & —     & —    & 68.8   \\
P-MIDN \cite{xu2021pyramidal}                 & —    & —    & —    & —    & —      & —    & —    & —    & —     & —    & —     & —    & —     & —     & —      & —     & —     & —    & —     & —    & 69.8   \\
\textit{Zhang} \cite{he2020momentum} & 86.3 & 72.9 & 71.2 & 59   & 36.3   & 80.2 & 84.4 & 75.6 & 30.8  & 83.6 & 53.2  & 75.1 & 82.7  & 87.1  & 37.7   & 54.6  & 74.2  & 59.1 & \textbf{79.8}  & 78.9 & 68.1   \\
\textbf{Ours}                  & 85.4 & 83.9 & 73.9 & \textbf{65.3} & \textbf{49.6}   & \textbf{86.8} & 83.3 & 75.3 & \textbf{45.4}  & 85.4 & \textbf{62.2}  & 74.9 & 84.5  & 88.2  & \textbf{56.9}   & 58.2  & 79.1  & 67.9 & 78.1  & 80.9 & \textbf{72.2}   \\ \hline
\multicolumn{20}{l}{\textit{Remark: “—” denotes not reported in the literature.}}\\
\end{tabular}
}
\end{table*}

\begin{figure*}[]
  \centering
   \includegraphics[width=0.9\linewidth]{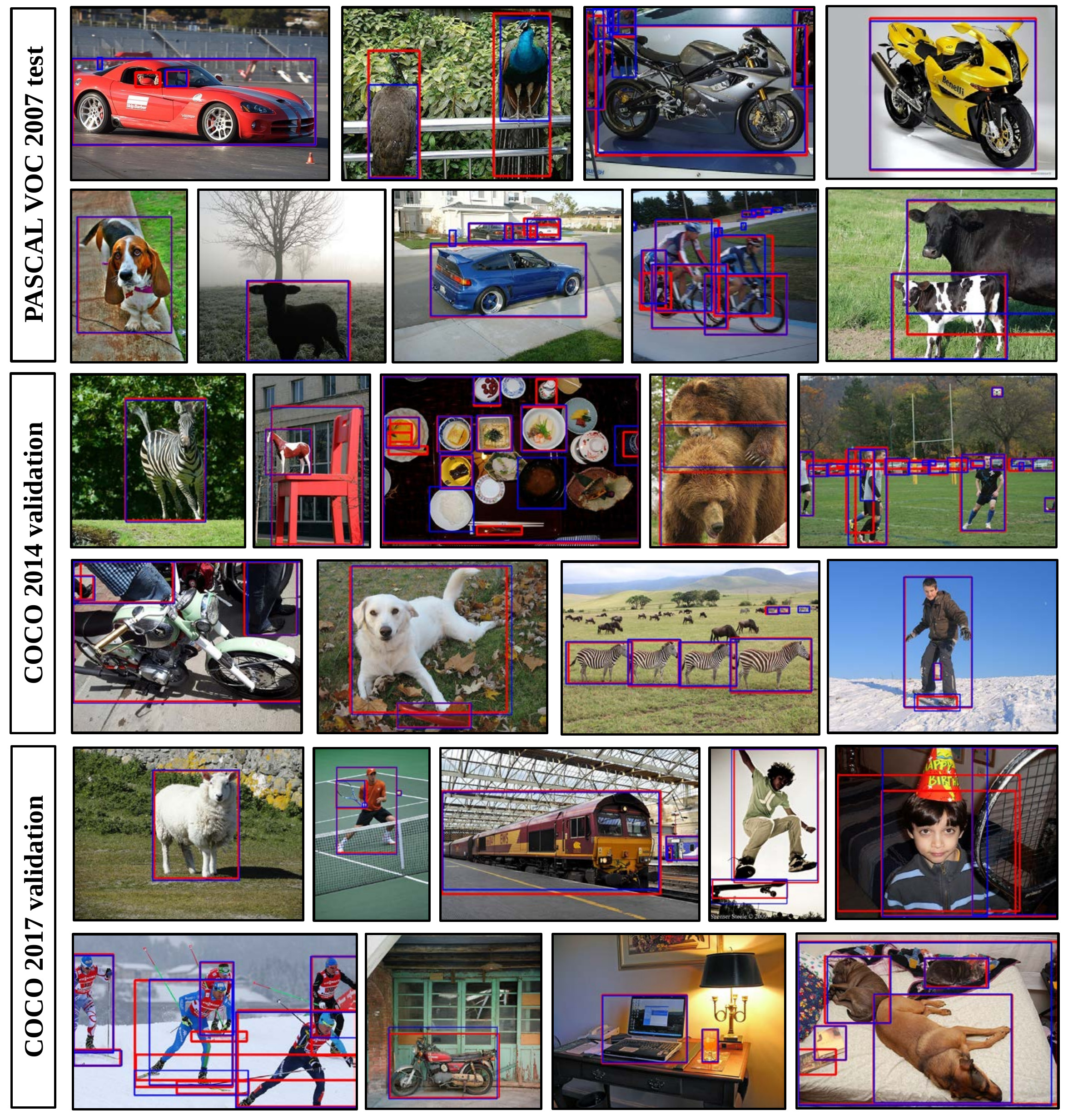}
   \caption{Some visual results of the detection obtained by JLWSOD on the PASCAL VOC 2007 test set, COCO 14 validation set, and COCO 17 validation set, where the blue boxes are the ground-truth boxes while red boxes are the predicted results.}
   \label{fig: Visualization results}
\end{figure*}

\textbf{Results on Pascal VOC 2007.} JLWSOD is compared against 9 state-of-the-art WSOD methods on the PASCAL VOC 2007 dataset. Following \cite{bilen2016weakly,tang2018pcl,zeng2019wsod2,xu2021pyramidal,tang2017multiple,li2019weakly,wan2019c,zhang2020weakly}, detection results are reported on the test set and localization results on trainval set, respectively. Table \ref{tab:map} reports the detection results on VOC 2007 where JLWSOD achieves 58.5\% mAP, surpassing all the compared 9 approaches. Remarkably, JLWSOD significantly outperforms previous WSOD methods on tiny objects, namely, bird, bottle, person, plant, by 3.3\%$\sim$22.1\%. Although JLWSOD cannot completely surpass the previous methods on some large objects, e.g., bike, sofa, etc., its mAP is very close to theirs. Table \ref{tab:corloc} reports the localization results on VOC 2007 where JLWSOD obtains distinguished performance gains on VOC 2007 trainval set, i.e., 5.2\% relative performance gain on average. The reason why the detection and localization results of JLWSOE sometimes drop on some categories, e.g., bike, sofa, etc., are discussed in Section \ref{sbusec:ablation}.

\begin{table}[ht]
  \caption{Comparison of AP performance (\%) on MS-COCO datasets.}
  \label{tab:coco}
  \centering
\begin{tabular}{lcc}
\hline
\multicolumn{1}{l}{Method} & \multicolumn{1}{c}{AP.50} & \multicolumn{1}{c}{AP {[}.50:.05:095{]}} \\ \hline
\multicolumn{3}{c}{MS-COCO 2014 validation set}                                                   \\ \hline
WSSDN \cite{bilen2016weakly}               & 19.2       & 9.5          \\
PCL \cite{tang2018pcl}                 & 19.4    & 9.2          \\
OICR \cite{tang2017multiple}     & 17.6     & 7.5   \\
WSOD$^2$ \cite{zeng2019wsod2}              & 22.7   & 10.8  \\
C-MIDN \cite{wan2019c}                    & 21.4  & 9.6    \\
Zhang et al. \cite{he2020momentum}      & 23.6     & 11.1   \\
P-MIDN   \cite{xu2021pyramidal}           & 27.7                      & 13.4                                     \\
\textbf{Ours}              & \textbf{33.9}             & \textbf{18.5}                            \\ \hline
\multicolumn{3}{c}{MS-COCO 2017 validation set}                                                   \\ \hline
WSSDN  \cite{bilen2016weakly}      & 19.2        & 9.5     \\
\textit{Zhang   et al.} \cite{he2020momentum}    & 23.8   & 11.2     \\
\textbf{Ours}              & \textbf{29.8}             & \textbf{14.9}                            \\ \hline
\multicolumn{3}{c}{MS-COCO 2017 test-dev set}                                                     \\ \hline
PCL\cite{tang2018pcl}                 & 19.4                      & 8.5                                      \\
\textit{Zhang et al.} \cite{he2020momentum}      & 24                        & 11.3                                     \\
\textbf{Ours}              & \textbf{29.7}             & \textbf{14.7}                            \\ \hline
\end{tabular}

\end{table}

\textbf{Results on MS-COCO.} Table \ref{tab:coco} summarizes the comparisons on the very challenging MS-COCO datasets, where the best performance of existing WSOD methods only achieves 27.7\%. Our proposed method still outperforms the most competitive P-MIDN \cite{xu2021pyramidal} and \textit{Zhang et al.} \cite{zhang2021weakly} up to 6\% in MS-COCO 2017.

Compared Table \ref{tab:coco} with Tables \ref{tab:map} and \ref{tab:corloc}, we can find that some results reflect that JLWSOD has a little weaker effect on PASCAL VOC 2007 dataset. This is because most of the images belonging to the PASCAL VOC 2007 dataset only contain one object category or one big object, while the images belonging to COCO datasets always contain multiple object categories. For this reason, the correlations between instance-wise or semantic-wise in COCO datasets are more abundant. Therefore, JLWSOD achieves significant improvement  on the COCO dataset than on PASCAL VOC 2007 dataset. Here, we give some visualized results obtained by our proposed method in Figure \ref{fig: Visualization results}.

\subsection{Ablation Study}
\label{sbusec:ablation}

\begin{table}[ht]
  \caption{Effectiveness of different modules in JLWSOD on VOC 2007.}
  \label{tab:ab4}
\centering
\begin{tabular}{ccccccc}
\hline
\begin{tabular}[c]{@{}c@{}}JLWSOD \\ Sub-Methods\end{tabular}
& M1 & M2& M3 & M4& mAP   &CorLoC \\ \hline
A  & $\surd$  &    &       &       & 45.5  & 59.6   \\
B  &    &  $\surd$ &    &  & 38.8  & 50.3   \\
C   &  $\surd$  & &  $\surd$      &   & 48.9  & 65.5   \\
D   &   & $\surd$   &$\surd$  &    & 41.1  & 51.1   \\
E    & $\surd$   &$\surd$  & $\surd$   &    & 52.5    & 67.5 \\
F   &  $\surd$   &  $\surd$     &     &  $\surd$     &\textbf{ 58.5} & \textbf{72.2} \\ \hline
\end{tabular}
\begin{flushleft}
\justify{Remark: The mAP (\%) mean the mean AP performance over all classes and CorLoc (\%) denotes the object localization performance. For brevity, we use M1, M2, M3, and M4 to indicate \textit{instance-wise detection branch}, \textit{semantic-wise prediction branch}, contrastive learning module, and \textit{interactive graph contrastive learning module}, respectively.}
\end{flushleft}
\end{table}

In this sub-section, a comprehensive study of JLWSOD is conducted through the ablation of different modules. For simplicity, we use M1, M2, M3, and M4 to respectively indicate the \textit{instance-wise detection branch}, \textit{semantic-wise prediction branch,} \textit{graph contrastive learning} (GCL), and \textit{interactive} GCL, generating 6 sub-methods (A to F). Note: M3 and M4 cannot be applied at the same time. Following the previous literature [20, 32, 34], the experiments are implemented on the VOC 2007 dataset. The characteristics of JLWSOD through ablation on four modules are investigated in Table \ref{tab:map} where mAP performance (\%) over all classes and CorLoc are reported. There are three significant observations:

\textbf{Each individual module is beneficial.} Comparing sub-methods C to F against A\&B (based on M1 or M2), we observe that each individual module is beneficial to the performance. However, the effect of M2 is far less than that of M1 because the semantic-wise prediction module (M2) has a higher demand for accurate labels than the object detection module (M1). To our best knowledge, the single-category nature of the majority of PASCAL VOC object images, resulting in weak correlations across different semantics, is the primary source of this issue. It significantly reduces the expected effectiveness of semantic-wise prediction branch (M2). Additionally, the majority of images in PASCAL VOC are foreground object images, i.e., a single item spans the majority of pixels, which causes the previous methods to locate only the incomplete object regions. Therefore, there are helpful instance-wise correlations between instances covering different parts of the target object, so the instance detection branch (M1) can obtain more excellent improvement in the overall performance. Of course, it is also possible that under weakly supervised environment, M2 alone cannot obtain enough information for training. Therefore, how to obtain accurate semantic labels under weak supervision is still a big challenge.

\textbf{The graph contrastive learning is effective.} Comparing A\&C, B\&D, we can see that graph contrastive learning can improve the performances (whether based on M1 or M2). Note that when M3 (C and D) is based on only one module (either M1 or M2), M3 is a standard GCL module as illustrated in Figure \ref{fig:iGCL} (a). The M3 in E is a simple multiplication-and-addition operation of two standard GCL, as shown in Figure \ref{fig:iGCL} (b).

\textbf{The interactive framework is important.} Comparing F against C to E, it is found that employing interactive framework iGCL achieves a large improvement (up to 17.4\% in mAP and 21.1\% in CorLoC). In addition, comparing E (M1+M2+M3) against F (M1+M2+M4), the effect of M3 is far less stable than M4.


\section{Conclusion and Future Work}
\label{sec:con}

In this paper, an interactive end-to-end WSOD framework called JLWSOD is presented which involves three novel modules: i) instance-wise detection branch, ii) semantic-wise prediction branch, iii) interactive graph contrastive (iGCL). The first two modules can learn the two types of WSOD-specific context information, i.e., the instance-wise correlation and the semantic-wise correlation via contrastive learning. Meanwhile, iGCL interactively incorporates two types of context information for \textit{joint} optimization. Hence, JLWSOD can be endowed with instance-level and semantic-level reasoning capabilities, which are beyond the exploration of many previous works. Extensive experiments verify the superiority of JLWSOD which improves 3.6\%$\sim$23.3\% on mAP and 3.4\%$\sim$19.7\% on CorLoc compared with other state-of-the-art methods. However, a drawback of our approach is that the precision of pseudo labels obtained by semantic-wise prediction branch is limited, which cannot sufficiently guide the subsequent WSOD task. Further explorations of semantic mechanisms under weakly supervised scenarios may further improve the performance of WSOD and are left as future work.



\bibliographystyle{IEEEtran}
\bibliography{IEEEabrv,example}

\vspace{11pt}

\vfill

\end{document}